\documentclass[11pt]{article}

\PassOptionsToPackage{hyperfootnotes=false}{hyperref}
\usepackage{acl}

\usepackage{times}
\usepackage{latexsym}
\usepackage[T1]{fontenc}
\usepackage[utf8]{inputenc}
\usepackage{microtype}
\usepackage{graphicx}
\usepackage{amsmath}
\usepackage{amssymb}
\usepackage{booktabs}
\usepackage{multirow}
\usepackage{xspace}
\usepackage{algorithm}
\usepackage{algorithmic}
\usepackage{subcaption}
\usepackage{xcolor}
\usepackage{colortbl}
\usepackage{tabularx}
\usepackage{fontawesome5}

\hypersetup{
  pdftitle={Omni-Prune: Query-Aware Unified Token Pruning for Efficient Omnimodal Large Language Models},
  pdfauthor={Yiming Zhong, Chang Nie, Caifeng Shan},
  pdfsubject={Training-free query-aware token pruning for omnimodal large language models}
}

\definecolor{oursrow}{RGB}{226, 235, 245}    
\definecolor{headerrow}{RGB}{200, 200, 200}  
\definecolor{fullrow}{RGB}{238, 238, 238}    

\newcommand{\eg}{e.g.,\xspace}
\newcommand{\ie}{i.e.,\xspace}

\title{Omni-Prune: Query-Aware Unified Token Pruning for Efficient \\Omnimodal Large Language Models}

\author{
  \textbf{Yiming Zhong}$^{1}$,\quad
  \textbf{Chang Nie}$^{1}$,\quad
  \textbf{Caifeng Shan}$^{1,*}$ \\[0.8ex]
  $^{1}$Nanjing University \\[0.6ex]
  {\small\texttt{zym@smail.nju.edu.cn}} \\[1.8ex]
  {\small
    \faGithub\enspace Code:\enspace
    \href{https://github.com/kimberlyii/Omni-Prune}
         {\textcolor{purple}{\texttt{github.com/kimberlyii/Omni-Prune}}}
  }
}

\begin{document}
\maketitle
\begingroup
\renewcommand{\thefootnote}{*}
\footnotetext{Corresponding author.}
\endgroup

\begin{abstract}
Omnimodal large language models (OmniLLMs) are rapidly extending multimodal reasoning to cover synchronized audio and video. However, the resulting audio-video token sequences are long, leading to high prefill latency and GPU memory usage at inference time. Existing token pruning methods, designed mainly for vision-only inputs, miss both the cross-modal links between audio and video and the user query that decides which content matters. To bridge this gap, we present \textbf{Omni-Prune}, a training-free, query-aware audio-visual token pruning framework that jointly removes redundancy from both modalities while keeping task-relevant cross-modal evidence. Specifically, Omni-Prune first splits the token sequence into adaptive time windows placed at audio saliency peaks, then scores audio and video tokens on a single scale that combines encoder attention with text-query relevance, and pairs related audio-video tokens so that they are kept together. Within each window, a final K-medoids step then selects a few representative tokens, adding diverse cues that score-based selection alone would miss.
Extensive experiments demonstrate that Omni-Prune outperforms established baseline methods, delivering up to \textbf{3.25$\times$} prefill speedup and \textbf{1.3$\times$} memory reduction while retaining over \textbf{99\%} of full-model performance.
\end{abstract}
\section{Introduction}
\label{sec:intro}

Recent omnimodal large language models (OmniLLMs)~\cite{xu2025qwen25omni, fu2025vita15, li2025baichuanomni15, wu2024nextgpt, lu2024unifiedio2} extend the multimodal paradigm beyond vision and language. They jointly ingest text, audio, and video within a single autoregressive backbone, and project all signals into a shared token space. This unified design unlocks fine-grained audiovisual reasoning, such as lip-synchronized dialogue understanding and sound-grounded event localization, that vision-only or audio-only systems cannot support.

\begin{figure}[t]
    \centering
    \includegraphics[width=0.98\columnwidth]{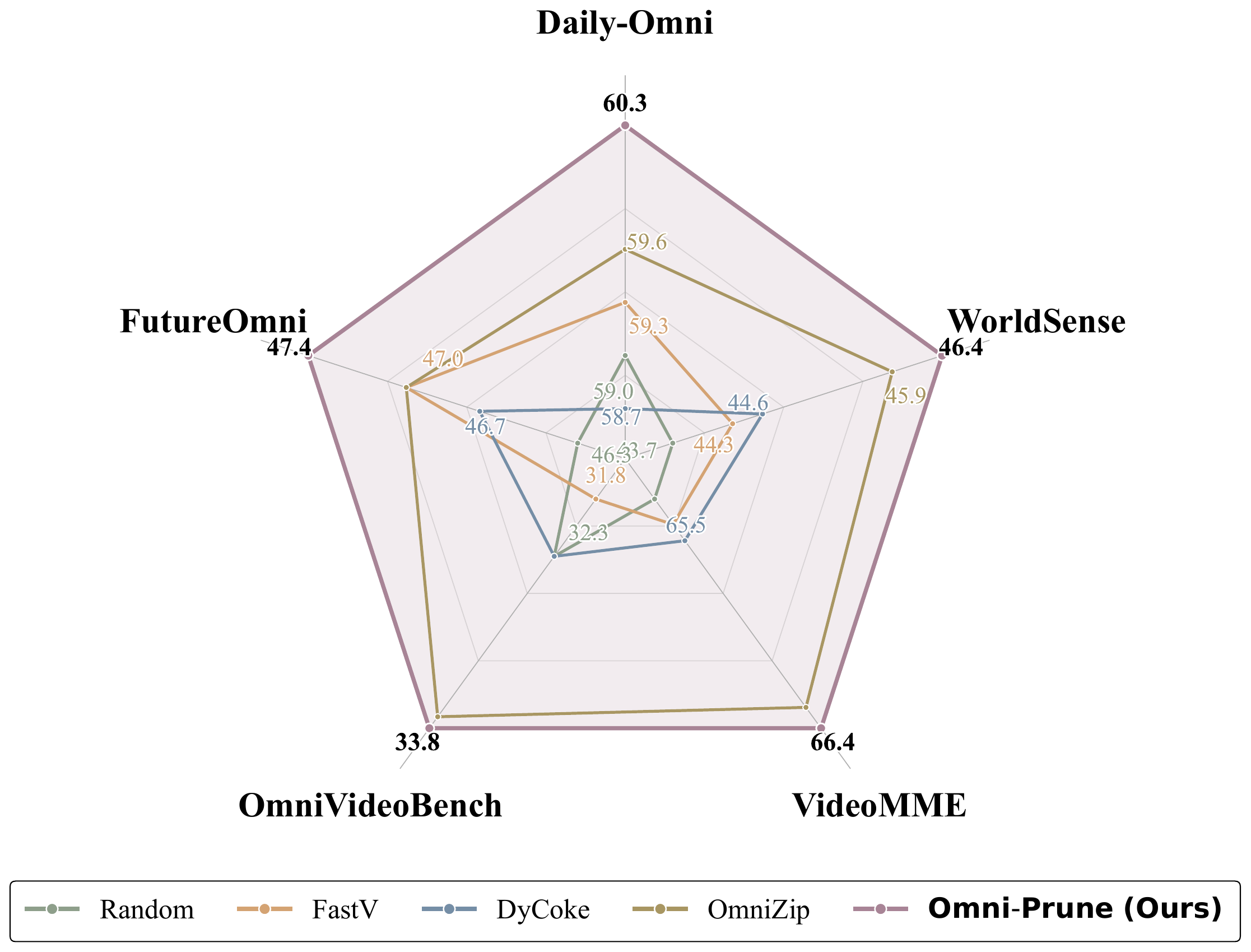}
    \caption{
\textbf{
Performance comparison on five omnimodal video benchmarks.}
Compared with other token compression baselines, our proposed Omni-Prune achieves the best performance with a matched or lower budget, and remains close to the full-token upper bound.}
    \label{fig:radar}
\end{figure}

Realizing this capability at scale, however, is bottlenecked by token volume. A one-minute clip at moderate sampling rates already produces on the order of $10^4$ interleaved audio-visual tokens~\cite{xu2025qwen25omni}. Self-attention scales quadratically with sequence length~\cite{vaswani2017attention}, so prefill latency and GPU memory quickly dominate inference cost. Token pruning is therefore a prerequisite for practical deployment.

\begin{figure*}[t]
    \centering
    \includegraphics[width=\textwidth]{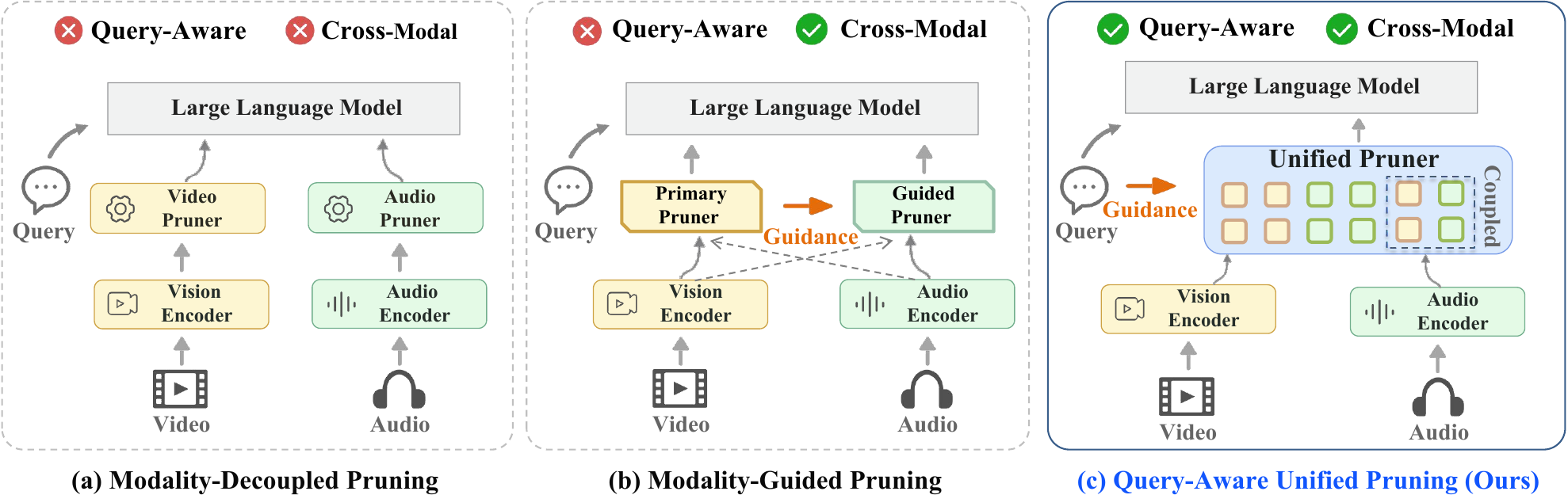}
    \caption{\textbf{Paradigms of token pruning methods.} (a) Modality-Decoupled Pruning: each modality is pruned independently, ignoring cross-modal redundancy and the user query. (b) Modality-Guided Pruning: a primary modality unilaterally drives the pruning of the other. (c) Query-Aware Unified Pruning (Ours): a query-conditioned pruner jointly scores and couples audio-video tokens.}
    \label{fig:teaser}
\end{figure*}

Token pruning has been extensively studied for image and video large language models~\cite{chen2024fastv, yang2024visionzip, xing2025pyramiddrop, tao2024dycoke, fan2026flashvid, zhang2024sparsevlm, zhang2025llavamini, shen2024longvu, ye2024vocollama}. Extending these methods to OmniLLMs is non-trivial, since audio carries complementary, temporally aligned evidence that a vision-only design cannot exploit~\cite{shi2022avhubert, sun2024videosalmonn}. Recent omnimodal pruning methods~\cite{tao2025omnizip, ding2026omnisift} take a step further~(Figure~\ref{fig:teaser} (a,b)), but they still fail to fully leverage all available modalities. The audio and video streams are processed in a sequential, one-way pipeline, where one modality unilaterally drives the reduction of the other rather than informing each other symmetrically. The text query, although central to the downstream task, is left out of the pruning decision entirely. So far, a token pruning approach that fully exploits all modalities has yet to be developed.

To this end, we propose \textbf{Omni-Prune}, a training-free token pruning framework for OmniLLMs~(Figure~\ref{fig:teaser}(c)) that introduces several technical innovations. First, we use audio saliency peaks as temporal boundaries to partition the token sequence into content-aware windows. Second, within each window, we integrate encoder attention with text-query relevance to score audio and video tokens on a unified scale, so that the resulting ranking reflects both modality importance and task intent. Third, building on these scores, we pair semantically correlated audio-video tokens and retain them jointly, ensuring that the pruning preserves cross-modal coherence rather than treating each modality in isolation.

Extensive experiments on five omnimodal benchmarks demonstrate that Omni-Prune consistently outperforms prior pruning methods while substantially reducing computational cost (Figure~\ref{fig:radar}). Our contributions in this work are as follows:

\begin{itemize}
    \item To the best of our knowledge, Omni-Prune is the first token pruning framework that incorporates all available modalities, including audio, video, and text, into the pruning decision.
    \item We design a unified scoring and cross-modal coupling mechanism that enables query-aware audio-video token selection within content-adaptive temporal windows.
    \item Evaluations on five benchmarks with 3B and 7B backbones show that Omni-Prune retains strong accuracy under aggressive token reduction while yielding notable savings in prefill latency and GPU memory consumption.
\end{itemize}

\section{Related Work}
\label{sec:related}

\paragraph{Omnimodal Large Language Models.}
The expansion of multimodal large language models toward richer perceptual channels has given rise to OmniLLMs, which receive text, image, video, and audio within a unified token sequence for end-to-end joint reasoning.
Compared with VideoLLMs~\cite{zhang2025videollama3, zhang2024llavavideo, bai2025qwen25vl, chen2024internvl25, li2024llavavid} that primarily target visual content understanding, OmniLLMs emphasize cross-modal collaborative perception, particularly the temporal alignment and semantic complementarity between audio and visual streams.
Qwen2.5-Omni~\cite{xu2025qwen25omni} achieves streaming comprehension through a Thinker-Talker architecture with TMRoPE-based temporal alignment, while VITA-1.5~\cite{fu2025vita15} introduces speech dialogue via multi-stage progressive training.
However, the massive number of audio and visual tokens in omnimodal inputs creates severe efficiency and memory bottlenecks that demand effective pruning.

\begin{figure*}[t]
    \centering
    \includegraphics[width=\textwidth]{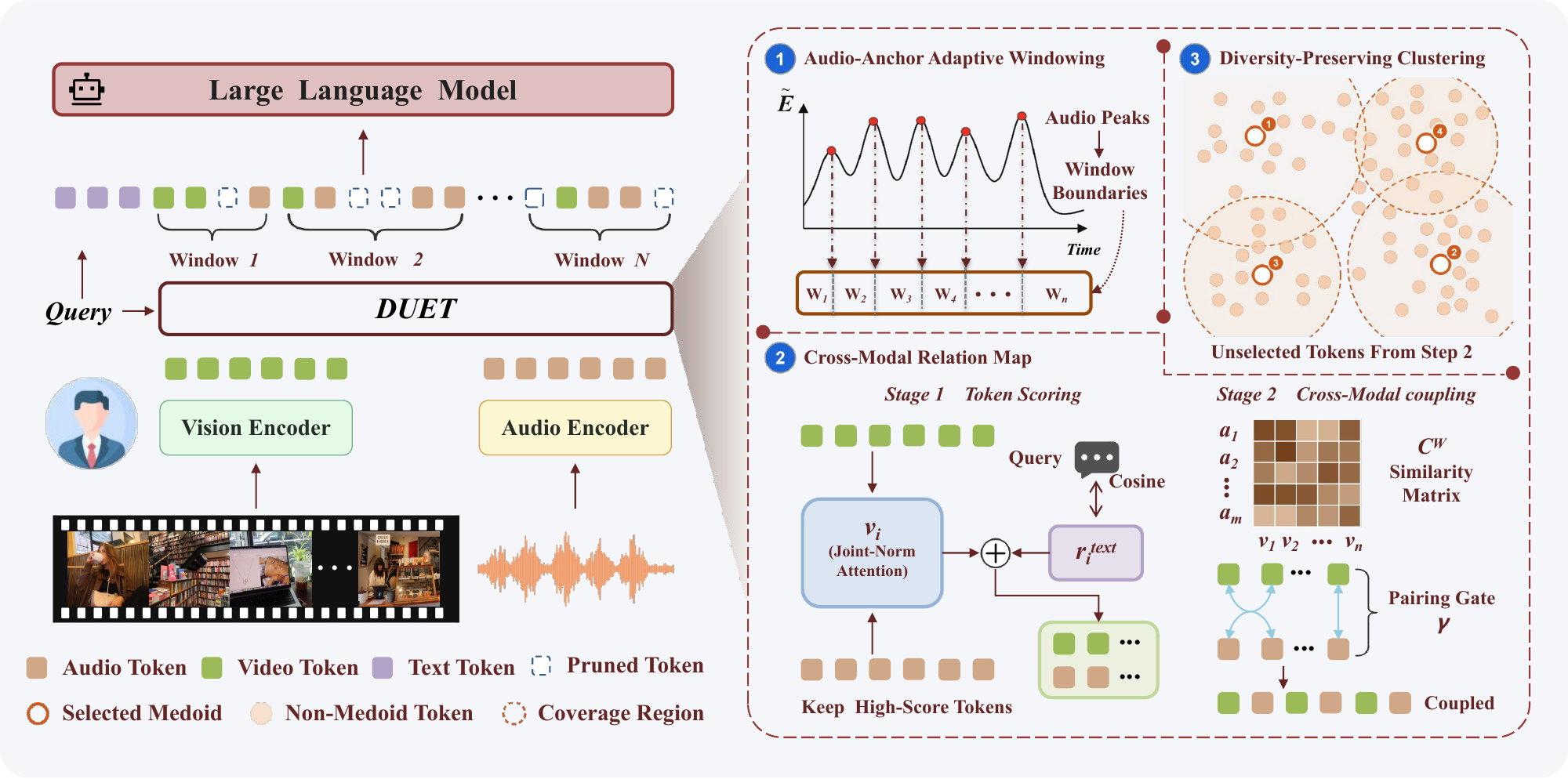}
    \caption{\textbf{Overview of the proposed Omni-Prune framework}. Given interleaved audio-video tokens from omnimodal encoders, the framework (1) segments the sequence into adaptive temporal windows anchored at audio saliency peaks, (2) scores and couples tokens via a cross-modal relation map, and (3) recovers representative residuals through K-medoids clustering. The pruned multimodal sequence is then passed to the LLM backbone for downstream reasoning.}
    \label{fig:method_overview}
\end{figure*}

\paragraph{Token Pruning.}
Existing token pruning methods have primarily been developed for multimodal models~\cite{chen2024fastv, shang2024llavaprumerge, yang2024visionzip, xing2025pyramiddrop, li2024tokenpacker, tao2024dycoke, li2024llavavid, fan2026flashvid}.
Representative approaches include attention-based pruning and merging~\cite{chen2024fastv, shang2024llavaprumerge}, entropy-guided token selection~\cite{yang2024visionzip}, layer-wise progressive dropping~\cite{xing2025pyramiddrop}, and spatiotemporal merging for video tokens~\cite{fan2026flashvid}.
For OmniLLMs, OmniZip~\cite{tao2025omnizip} uses audio attention as anchors to allocate video compression budgets. Conversely, OmniSIFT~\cite{ding2026omnisift} reverses this flow by pruning video first and then guiding audio retention with visual semantics.

\section{Method}
\label{sec:method}

In this section, we present Omni-Prune, a training-free cross-modal token pruning framework for OmniLLMs, illustrated in Figure~\ref{fig:method_overview}. The framework is built around two core modules: \emph{Audio-Anchor Adaptive Windowing} (AAW), which segments the token sequence into content-aware temporal windows aligned with audio-salient events, and \emph{Cross-Modal Relation Map} (CMRM), which jointly scores audio and video tokens within each window and identifies cross-modal merge pairs. To further preserve representational diversity, we employ K-medoids clustering to recover a compact set of representative tokens.

\subsection{Background}
\label{sec:prelim}

A typical OmniLLM follows an encoder-decoder architecture: modality-specific encoders first convert raw video and audio into token embeddings, which are then interleaved with text tokens and fed into a shared LLM decoder for autoregressive generation. Specifically, the video is sampled into $F$ frames $\{I_1, \ldots, I_F\}$, and each modality is encoded via dedicated encoders $\mathcal{F}_{\text{vis}}$, $\mathcal{F}_{\text{aud}}$:
\begin{align}
    \mathbf{Z}_{\text{vis}} &= \mathcal{F}_{\text{vis}}(I_1, \ldots, I_F) \in \mathbb{R}^{N_v \times D}, \\
    \mathbf{Z}_{\text{aud}} &= \mathcal{F}_{\text{aud}}(\mathbf{w}) \in \mathbb{R}^{N_a \times D},
\end{align}
where $\mathbf{w}$ denotes the raw audio waveform, $N_v$/$N_a$ are the numbers of visual/audio tokens, and $D$ is the hidden dimension.

These token sequences are interleaved with text tokens by temporal order and projected into a shared embedding space, producing a flat sequence $\mathbf{X} = [\mathbf{x}_1, \ldots, \mathbf{x}_L] \in \mathbb{R}^{L \times D}$, where each token $\mathbf{x}_i$ belongs to one of three modalities: video ($\mathcal{V}$), audio ($\mathcal{A}$), or text ($\mathcal{T}$). The encoders additionally yield attention scores $\mathbf{s}^v \in \mathbb{R}^{|\mathcal{V}|}$ and $\mathbf{s}^a \in \mathbb{R}^{|\mathcal{A}|}$, which reflect the encoder-level saliency of each token. Given a keep ratio $\alpha \in (0,1]$, Omni-Prune selects at most $B = \lfloor \alpha \cdot (|\mathcal{A}| + |\mathcal{V}|) \rceil$ audio-visual tokens (the \emph{total budget}) to retain while keeping all text tokens intact.

\subsection{Audio-Anchor Adaptive Windowing}
\label{sec:windowing}

Prior pruning approaches typically operate on fixed temporal windows~\cite{tao2025omnizip, ding2026omnisift} or process the entire sequence globally~\cite{chen2024fastv}, ignoring that semantically meaningful events are unevenly distributed over time. We observe that audio-salient events (speech onsets, sound effects, music transitions) frequently co-occur with important video moments, making audio attention peaks natural segment boundaries.

\paragraph{Audio Saliency Computation.} To bridge these two temporal granularities, we uniformly partition the audio token sequence into $F$ segments, each temporally aligned with one video frame. The audio saliency of the $f$-th segment is then defined as the mean encoder attention score over its constituent tokens:
\begin{equation}
    E_f = \frac{1}{|T_f|} \sum_{i \in T_f} s^a_i, \quad f = 1, \ldots, F,
\end{equation}
where $T_f$ denotes the set of audio token indices assigned to the $f$-th segment. This per-segment energy reflects how much the audio encoder attends to each temporal region. The raw energy curve is subsequently smoothed via 1D average pooling with kernel size 3 to suppress transient fluctuations, yielding the smoothed saliency $\tilde{E}$.

\begin{table*}[htbp]
\centering
\small
\caption{\textbf{Performance of different token pruning methods on omnimodal QA benchmarks.} \textbf{Bold} indicates the best result and \underline{underline} the second best. The ``-'' symbol indicates an Out-of-Memory (OOM) failure, and such entries are excluded from the average.}
\label{tab:main}
\resizebox{\linewidth}{!}{%
\begin{tabular}{l c c c c c c c}
\toprule
\textbf{Method} & \textbf{Retained} & \textbf{Daily-Omni} & \textbf{WorldSense} & \textbf{VideoMME} & \textbf{OmniVideoBench} & \textbf{FutureOmni} & \textbf{Avg.} \\
\midrule
\rowcolor{headerrow}
\multicolumn{8}{c}{\textit{Qwen2.5-Omni-7B}} \\
\midrule
\rowcolor{fullrow}
Full Tokens & 100\% & 62.5 & 46.8 & 66.0 & 33.8 & 47.5 & 100\% \\
\midrule
Random & 55\% & 59.0 & 43.7 & 65.3 & 32.3 & 46.3 & 96.0\% \\
FastV & 50\% & 59.3 & 44.3 & -- & 31.8 & \underline{47.0} & 95.6\% \\
DyCoke & 50\% & 58.7 & 44.6 & 65.5 & 32.3 & 46.7 & 96.5\% \\
OmniZip & 45\% & \underline{59.6} & \underline{45.9} & \underline{66.3} & \underline{33.7} & \underline{47.0} & \underline{98.5\%} \\
\rowcolor{oursrow}
\textbf{Omni-Prune (Ours)} & 45\% & \textbf{60.3} & \textbf{46.4} & \textbf{66.4} & \textbf{33.8} & \textbf{47.4} & \textbf{99.2\%} \\
\midrule
Random & 40\% & 58.1 & 42.9 & 64.8 & 31.1 & 45.8 & 94.2\% \\
FastV & 35\% & 58.7 & 43.6 & -- & 31.7 & \underline{46.4} & 94.6\% \\
DyCoke & 35\% & 57.8 & 43.8 & 65.2 & \underline{31.9} & 46.2 & 95.3\% \\
OmniZip & 35\% & \underline{58.8} & \underline{45.3} & \underline{66.1} & \textbf{34.3} & \textbf{46.5} & \underline{98.1\%} \\
\rowcolor{oursrow}
\textbf{Omni-Prune (Ours)} & 35\% & \textbf{60.4} & \textbf{46.2} & \textbf{66.3} & \textbf{34.3} & \textbf{46.5} & \textbf{99.0\%} \\
\midrule
\rowcolor{headerrow}
\multicolumn{8}{c}{\textit{Qwen2.5-Omni-3B}} \\
\midrule
\rowcolor{fullrow}
Full Tokens & 100\% & 61.9 & 46.4 & 62.8 & 33.6 & 46.5 & 100\% \\
\midrule
Random & 55\% & 56.9 & 43.1 & 60.9 & 31.2 & 46.1 & 94.8\% \\
FastV & 50\% & \underline{58.0} & 44.4 & -- & 30.6 & 46.5 & 95.1\% \\
DyCoke & 50\% & 57.5 & 44.0 & 61.6 & 31.0 & 46.0 & 95.4\% \\
OmniZip & 45\% & \underline{58.0} & \underline{45.2} & \underline{62.8} & \underline{32.4} & \underline{46.8} & \underline{97.6\%} \\
\rowcolor{oursrow}
\textbf{Omni-Prune (Ours)} & 45\% & \textbf{58.8} & \textbf{45.9} & \textbf{63.3} & \textbf{32.6} & \textbf{47.0} & \textbf{98.6\%} \\
\midrule
Random & 40\% & 55.8 & 42.2 & 60.3 & 30.8 & 45.2 & 93.2\% \\
FastV & 35\% & \underline{57.2} & 43.7 & -- & 30.1 & 46.1 & 93.8\% \\
DyCoke & 35\% & 56.2 & 43.2 & 61.0 & 30.6 & 45.4 & 93.9\% \\
OmniZip & 35\% & 56.8 & \underline{44.3} & \underline{62.7} & \textbf{32.0} & \underline{46.5} & \underline{96.5\%} \\
\rowcolor{oursrow}
\textbf{Omni-Prune (Ours)} & 35\% & \textbf{58.5} & \textbf{45.4} & \textbf{62.9} & \underline{31.6} & \textbf{46.9} & \textbf{97.5\%} \\
\bottomrule
\end{tabular}%
}
\end{table*}

\paragraph{Peak-Guided Segmentation.} Local peaks in $\tilde{E}$ (\ie $\tilde{E}_f \geq \tilde{E}_{f-1}$ and $\tilde{E}_f \geq \tilde{E}_{f+1}$) serve as candidate window boundaries. Since they correspond to audio-salient time points where cross-modal information density is highest. To balance granularity and context, consecutive boundaries are constrained to a minimum gap of $f_{\min} = 2$, which suppresses spurious noisy peaks and avoids fragmenting a coherent event.

\subsection{Cross-Modal Relation Map}
\label{sec:relation}

Once the token sequence is partitioned into adaptive windows, Omni-Prune scores each audio and video token within a window to determine its retention priority. Since the two modalities originate from separate encoders, we construct the Cross-Modal Relation Map (CMRM) that fuses attention saliency with text-query relevance into a unified scoring framework.
\paragraph{Token Importance Scoring.} Within each window $w$, we jointly evaluate the importance of audio and video tokens. The encoder attention score of each token $i$ is derived from the last self-attention layer by averaging across all heads and summing over all query positions:
\begin{equation}
    s_i = \sum_{j} \frac{1}{H} \sum_{h=1}^{H} \text{softmax}\!\left(\frac{\mathbf{q}_j^{(h)} \cdot \mathbf{k}_i^{(h)}}{\sqrt{d_k}}\right),
\end{equation}
where $H$ is the number of attention heads, $\mathbf{q}_j^{(h)}$ and $\mathbf{k}_i^{(h)}$ are the query of the $j$-th token and key of the $i$-th token at head $h$, respectively. Joint min-max normalization is then applied across both modalities within the window:
\begin{equation}
    v_i = \frac{s^w_i - \min(\mathbf{s}^w)}{\max(\mathbf{s}^w) - \min(\mathbf{s}^w) + \epsilon}, \quad \epsilon = 10^{-6}.
\end{equation}
By normalizing audio and video scores together rather than independently, this places both modalities on a shared $[0, 1]$ scale and enables direct cross-modal comparison within each window.

Encoder attention alone captures modality-internal saliency but is agnostic to the user's intent. To incorporate task awareness, we measure how semantically close each multimodal token is to the user query. The text-query relevance is defined as:
\begin{equation}
    r_i^{\text{text}} = \frac{1}{2}\left(\max_j\; \frac{\mathbf{x}_i^\top \mathbf{t}_j}{\|\mathbf{x}_i\| \cdot \|\mathbf{t}_j\|} + 1\right),
\end{equation}
where $\mathbf{x}_i$ is the embedding of the $i$-th multimodal token and $\mathbf{t}_j$ is the $j$-th text token from the user query. The cosine similarity $\in [-1, 1]$ is rescaled to $[0, 1]$ via the affine mapping. The final composite score integrates both signals:
\begin{equation}
    \phi_i = v_i + r_i^{\text{text}}.
\end{equation}
Tokens whose normalized value $v_i$ exceeds the per-window average $\bar{v}^w$ or whose text relevance $r_i^{\text{text}}$ exceeds $\max(\bar{r}^{\text{text}}, \tau)$ are designated as \emph{keep candidates}, where $\tau$ serves as a floor threshold corresponding to marginal semantic relevance. Within each window, candidates are ranked by $\phi_i$ and at most $B_{\text{keep}}^w = \lfloor \alpha \cdot \rho_{\text{keep}} \cdot (|\mathcal{A}^w| + |\mathcal{V}^w|) \rceil$ are selected, where $\rho_{\text{keep}}$ caps the keep fraction to leave room for cross-modal paired retention.

\paragraph{Cross-Modal Coupling.} The composite scoring identifies important tokens independently. To further exploit the cross-modal structure of omnimodal inputs, we observe that certain audio and video tokens represent the same underlying event and should be retained as a pair to preserve coherence. We thus compute the audio-video similarity within each window:
\begin{equation}
    C^w_{ij} = \frac{\mathbf{a}_i^\top \mathbf{v}_j}{\|\mathbf{a}_i\| \cdot \|\mathbf{v}_j\|}, \quad \mathbf{C}^w \in [-1, 1]^{|\mathcal{A}^w| \times |\mathcal{V}^w|},
\end{equation}
where $\mathbf{a}_i$ and $\mathbf{v}_j$ are the embeddings of the $i$-th audio token and $j$-th video token within window $w$. Each entry $C^w_{ij}$ measures the cosine similarity between an audio-video token pair. For unselected tokens, bidirectional greedy matching pairs each audio token with its most similar video counterpart using normalized similarity $\text{sim}(a_i, v_j) = (C^w_{ij} + 1)/2$. Pairs exceeding a gate threshold $\gamma$ are jointly retained. Any remaining budget is then filled from unselected tokens sorted by $\phi_i$, with a preference for the minority modality so as to maintain cross-modal balance.

\subsection{Diversity-Preserving Clustering}
\label{sec:clustering}

The preceding stages retain the most important and semantically coupled tokens. However, the unselected tokens are not uniformly uninformative, as some carry complementary details such as ambient sounds, background objects, and peripheral scene context that would otherwise be lost entirely. Discarding them outright risks collapsing the retained set onto a narrow region of the embedding space dominated by high-scoring events, which is harmful for long-horizon or scene-heavy queries where minor cues can disambiguate the answer. We therefore reintroduce a small fraction of these tokens chosen to maximize representational coverage rather than per-token importance, providing the LLM with a coarse-grained backdrop.

To this end, Omni-Prune applies K-medoids clustering~\cite{park2024kmedoids} independently within each window and modality. We choose K-medoids over K-means~\cite{macqueen1967kmeans} because each medoid is itself a real token already produced by the encoder, whereas a K-means centroid is an averaged vector and would therefore be out-of-distribution for the downstream LLM. For each group of unselected tokens, a fraction $\rho_c$ of them are selected as medoids via farthest-first traversal in cosine distance space to maximize representational spread, a greedy procedure that runs in $O(k \cdot n)$ per window and adds negligible overhead. Operating per window and per modality avoids spurious clusters that would otherwise span unrelated temporal segments or mix heterogeneous modality statistics. Finally, all remaining tokens are discarded from the sequence.
\section{Experiments}
\label{sec:experiments}

\subsection{Experimental Setting}
\label{sec:setup}

\paragraph{Benchmarks.} We evaluate Omni-Prune on five omnimodal video understanding benchmarks that together cover both general-purpose and audio-centric reasoning. \textbf{VideoMME}~\cite{fu2024videomme} provides a comprehensive multi-domain testbed, with short, medium, and long clips paired with synchronized audio tracks. Centered on everyday scenes, \textbf{Daily-Omni}~\cite{zhou2025dailyomni} requires models to jointly reason over speech, ambient sound, and visual context. \textbf{WorldSense}~\cite{hong2025worldsense} further stresses real-world knowledge grounding by sourcing diverse audio-visual events at scale. \textbf{OmniVideoBench}~\cite{li2025omnivideobench} is curated specifically to probe fine-grained omnimodal understanding. Finally, \textbf{FutureOmni}~\cite{chen2026futureomni} targets long-horizon, future-oriented reasoning over synchronized audio-visual streams.

\paragraph{Baselines.} We compare Omni-Prune against several state-of-the-art (SOTA) token pruning approaches. \textbf{OmniZip}~\cite{tao2025omnizip} leverages audio attention for video token pruning. \textbf{FastV}~\cite{chen2024fastv} prunes visual tokens in a layer-wise manner, while \textbf{DyCoke}~\cite{tao2024dycoke} reduces temporal redundancy by merging video tokens across frames and dynamically evicting KV cache entries during decoding. We also include a random pruning baseline for rigorous comparison.

\paragraph{Implementation Details.} Omni-Prune is built on the Qwen2.5-Omni (7B and 3B) and runs on NVIDIA A6000 (48GB) GPUs, with FlashAttention~\cite{dao2022flashattention} enabled throughout to reduce memory consumption. For hyperparameter settings, we set the text-relevance floor $\tau = 0.55$, the cross-modal pairing gate $\gamma = 0.6$, and the residual cluster ratio $\rho_c = 0.1$. 
In addition, we set $\alpha$ to $0.4$ and $0.28$, corresponding to retention rates of $45\%$ and $35\%$, respectively.

\subsection{Main Results}
\label{sec:main_results}

We follow a unified evaluation protocol for every token pruning method in Table~\ref{tab:main}, keeping the decoding configuration, prompt template, and input pipeline identical. Each method is run at two distinct retention rates, covering both a moderate and an aggressive budget. To make the benchmarks comparable despite their different absolute scales, we normalize each entry by the full-token score on the same benchmark and report Avg.\ as the mean of them, with the full-token model anchored at $100\%$.

\paragraph{Comparison with SOTAs.} The headline takeaway from Table~\ref{tab:main} is that Omni-Prune is essentially lossless on the 7B backbone. At the moderate $45\%$ budget it retains $99.2\%$ of full-model accuracy and ranks first on every benchmark, and even after discarding roughly two-thirds of the input tokens at the aggressive $35\%$ budget it still holds $99.0\%$. The competing strategies fall short, and the way they fall short is informative. As a sanity check, the random baseline trails our method by a clear margin even when given a more generous budget, confirming that \emph{which} tokens are retained matters more than \emph{how many} are kept. Specifically, FastV scores tokens from attention magnitudes alone, a single-criterion design whose quadratic cost makes it prohibitive on long interleaved audio-video sequences and triggers out-of-memory failures on VideoMME. DyCoke targets temporal redundancy by merging similar tokens across consecutive video frames and evicting KV cache entries during decoding. The merging is locality-based and assumes that adjacent frames are near-duplicates, a prior that holds for static scenes but breaks down on audio-driven benchmarks, where the discriminative content shifts on the audio cadence. By contrast, OmniZip is the most competitive baseline and retains its second-place ranking, but its audio-to-video saliency rule is unidirectional and query-agnostic, so a silent yet visually critical token is easily down-weighted by its loud neighbor while a loud yet off-query token is kept.

\begin{table}[t!]
\centering
\small
\caption{\textbf{Efficiency comparison on Daily-Omni.} Peak GPU memory and prefill time are measured on a single A6000 GPU with Qwen2.5-Omni-7B and 3B. \textbf{Bold} and \underline{underline} values indicate the best and second-best results among the pruning methods, respectively.}
\label{tab:efficiency}
\resizebox{\linewidth}{!}{%
\begin{tabular}{lcc}
\toprule
\textbf{Method} & \textbf{GPU Mem.}$\downarrow$ & \textbf{Prefill Time}$\downarrow$ \\
\midrule
\rowcolor[RGB]{225,225,225}
\multicolumn{3}{c}{\textit{Qwen2.5-Omni-7B}} \\
\midrule
\rowcolor[RGB]{247,247,247}
Full Tokens (100\%) & 28.7G & 899ms (1.00$\times$) \\
DyCoke (50\%) & 22.9G~\textcolor{red}{\scriptsize($\Downarrow$5.8)} & 403ms (2.23$\times$) \\
OmniZip (45\%) & \underline{22.5G}~\textcolor{red}{\scriptsize($\Downarrow$6.2)} & 373ms (2.41$\times$) \\
\rowcolor[RGB]{240,247,253}
Omni-Prune (45\%) & \underline{22.5G}~\textcolor{red}{\scriptsize($\Downarrow$6.2)} & \underline{345ms} (\underline{2.61$\times$}) \\
\rowcolor[RGB]{240,247,253}
Omni-Prune (35\%) & \textbf{22.3G}~\textcolor{red}{\scriptsize($\Downarrow$6.4)} & \textbf{277ms} (\textbf{3.25$\times$}) \\
\midrule
\rowcolor[RGB]{225,225,225}
\multicolumn{3}{c}{\textit{Qwen2.5-Omni-3B}} \\
\midrule
\rowcolor[RGB]{247,247,247}
Full Tokens (100\%) & 18.4G & 567ms (1.00$\times$) \\
DyCoke (50\%) & 13.0G~\textcolor{red}{\scriptsize($\Downarrow$5.4)} & 209ms (2.71$\times$) \\
OmniZip (45\%) & \textbf{12.8G}~\textcolor{red}{\scriptsize($\Downarrow$5.6)} & 223ms (2.54$\times$) \\
\rowcolor[RGB]{240,247,253}
Omni-Prune (45\%) & \underline{12.9G}~\textcolor{red}{\scriptsize($\Downarrow$5.5)} & \underline{175ms} (\underline{3.24$\times$}) \\
\rowcolor[RGB]{240,247,253}
Omni-Prune (35\%) & \textbf{12.8G}~\textcolor{red}{\scriptsize($\Downarrow$5.6)} & \textbf{156ms} (\textbf{3.63$\times$}) \\
\bottomrule
\end{tabular}%
}
\end{table}

\paragraph{Behavior Across Model Scales.} The 3B backbone is uniformly more fragile to pruning than the 7B one, and the methods separate more clearly at the smaller scale. On the 7B backbone the methods are bunched within a few points of full accuracy; on the 3B backbone the gap between Omni-Prune and the other methods steadily widens, especially under the aggressive $35\%$ budget. Crucially, this gap grows rather than shrinks as model capacity decreases, indicating that the advantage stems from genuinely better token selection rather than from the larger backbone having enough spare capacity to mask other methods' pruning errors.

\begin{table*}[t]
\centering
\setlength{\tabcolsep}{4.2pt}
\begin{tabular}{lccccc}
\toprule
Variant & Daily-Omni & WorldSense & VideoMME & OmniVideoBench & FutureOmni \\
\midrule
\rowcolor{oursrow}
Full            & 60.3 & 46.4 & 66.4 & 33.8 & 47.4 \\
w/o Clustering  & 60.1~\textcolor{red}{\scriptsize($-$0.2)} & 46.1~\textcolor{red}{\scriptsize($-$0.3)} & 66.1~\textcolor{red}{\scriptsize($-$0.3)} & 33.4~\textcolor{red}{\scriptsize($-$0.4)} & 47.2~\textcolor{red}{\scriptsize($-$0.2)} \\
\midrule
\rowcolor{oursrow}
Full            & 60.4 & 46.2 & 66.3 & 34.3 & 46.5 \\
w/o Clustering  & 59.2~\textcolor{red}{\scriptsize($-$1.2)} & 46.0~\textcolor{red}{\scriptsize($-$0.2)} & 64.5~\textcolor{red}{\scriptsize($-$1.8)} & 34.1~\textcolor{red}{\scriptsize($-$0.2)} & 46.4~\textcolor{red}{\scriptsize($-$0.1)} \\
\bottomrule
\end{tabular}
\caption{\textbf{Effect of diversity-preserving clustering on Qwen2.5-Omni-7B.} We compare Omni-Prune with and without the residual K-medoids step on five benchmarks, and report per-benchmark scores together with the absolute drop. The top block corresponds to the $45\%$ budget and the bottom block to the $35\%$ budget.}
\label{tab:ablation_cluster}
\end{table*}

\paragraph{Fine-Grained Analysis.} A striking observation in Table~\ref{tab:main} is that Omni-Prune actually \emph{exceeds} the full-token baseline on several benchmarks, most notably \emph{VideoMME}, \emph{OmniVideoBench}, and \emph{FutureOmni}. These benchmarks tend to contain a substantial amount of redundant or off-topic context, where pruning can act as a mild denoiser that helps surface the answer-bearing signal.

\subsection{Efficiency Analyses}
\label{sec:efficiency}

Beyond accuracy, practical deployment of OmniLLMs demands low latency and manageable memory footprint.
Table~\ref{tab:efficiency} reports the prefill time and peak GPU memory on Daily-Omni.
On the 7B model, Omni-Prune at 35\% retention reduces prefill time from 899\,ms to 277\,ms (3.25$\times$) and trims peak memory by 6.4\,G. On the 3B backbone the gap widens further, reaching 3.63$\times$ acceleration with 5.6\,G less memory. Notably, DyCoke keeps 50\% of tokens and OmniZip keeps 45\%, yet Omni-Prune still achieves the shortest prefill time under a more aggressive pruning ratio.

\subsection{Ablation Study}
\label{sec:ablation}

\begin{table}[t!]
\centering
\small
\caption{\textbf{Ablation study of CMRM on Qwen2.5-Omni-7B at the $45\%$ budget.} To verify the contribution of each design choice in CMRM, we run an ablation that measures how the two core ingredients shape final accuracy: token importance scoring (Scoring) and cross-modal coupling (Coupling).}
\label{tab:ablation}
\resizebox{\linewidth}{!}{%
\begin{tabular}{cc|ccc}
\toprule
\multicolumn{2}{c|}{\textbf{Settings}} & \multicolumn{3}{c}{\textbf{Benchmark}} \\
\cmidrule(lr){1-2} \cmidrule(lr){3-5}
Scoring & Coupling & Daily-Omni & OmniV. & FutureO. \\
\midrule
\rowcolor{oursrow}
\textcolor{green!60!black}{$\checkmark$} & \textcolor{green!60!black}{$\checkmark$} & 60.3 & 33.8 & 47.4 \\
\textcolor{red}{$\times$}     & \textcolor{green!60!black}{$\checkmark$} & 59.2~\textcolor{red}{\scriptsize($-$1.1)} & 33.1~\textcolor{red}{\scriptsize($-$0.7)} & 46.5~\textcolor{red}{\scriptsize($-$0.9)} \\
\textcolor{green!60!black}{$\checkmark$} & \textcolor{red}{$\times$}     & 59.7~\textcolor{red}{\scriptsize($-$0.6)} & 33.6~\textcolor{red}{\scriptsize($-$0.2)} & 47.1~\textcolor{red}{\scriptsize($-$0.3)} \\
\bottomrule
\end{tabular}%
}
\end{table}

\begin{figure}[t]
\centering
\includegraphics[width=\linewidth]{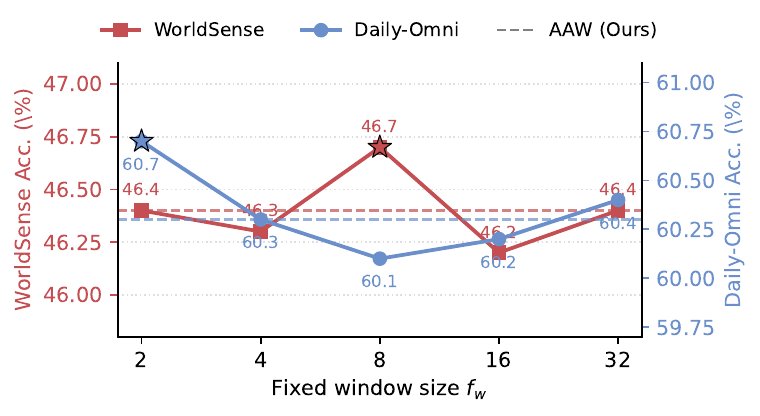}
\caption{\textbf{Ablation study of AAW.} We sweep the fixed window size $f_w$ under the $45\%$ retention budget on Qwen2.5-Omni-7B and compare against our audio-anchor adaptive windows, on both WorldSense (left axis) and Daily-Omni (right axis).}
\label{fig:aaw_curve}
\end{figure}

\paragraph{Ablation Study on AAW.} We examine the role of content-aware windowing by sweeping the fixed window size $f_w$ on WorldSense and Daily-Omni and comparing each setting against our audio-anchor adaptive windows (Figure~\ref{fig:aaw_curve}). The fixed-window curves are non-monotonic and noticeably sensitive to $f_w$ on both benchmarks, but the best operating point differs: WorldSense peaks at $f_w{=}8$ while Daily-Omni peaks at $f_w{=}2$, and either choice is suboptimal on the other benchmark. With a fixed stride the boundaries are content-agnostic and frequently cut through ongoing speech events, so the resulting windows mix unrelated material and the per-window token scoring becomes less stable. The optimal $f_w$ also depends on the cadence of each dataset, which is why no single fixed choice dominates across the sweep. AAW instead places boundaries at audio saliency peaks, so each window stays content-coherent and adapts to the actual event rhythm of the input. As a result, AAW lands close to the best fixed setting on both datasets without any per-task tuning, confirming that audio-anchored adaptive windowing is a more reliable default than any fixed stride.

\paragraph{Ablation Study on CMRM.} Table~\ref{tab:ablation} reports the ablation study that isolates the two key ingredients of CMRM, namely the token importance scoring and the cross-modal coupling. Disabling the scoring stage tends to incur the larger drop on the three benchmarks. Without a dedicated importance signal, every token is treated as roughly equally worth keeping, and the budget gets spread evenly rather than concentrated on the few tokens that actually carry the answer. As a result, informative content is more easily crowded out by redundant background. Removing the coupling step is less damaging in isolation but still leaves a visible gap, because per-modality ranking can independently keep an audio token while dropping its visually grounded counterpart (or vice versa), breaking pairs that are only useful when retained jointly. Overall, the two ingredients are complementary, with scoring deciding which individual tokens deserve the budget and coupling preserving the cross-modal pairs.

\paragraph{Ablation Study on Diversity-Preserving Clustering.} As shown in Table~\ref{tab:ablation_cluster}, we evaluate the effect of the residual K-medoids step. Primarily, clustering yields consistent gains, and its benefit scales with how aggressive the pruning is. At the looser $45\%$ budget the drops are uniformly mild, whereas at the tighter $35\%$ budget the gap widens markedly, most notably on VideoMME and Daily-Omni. A plausible explanation is that these two benchmarks tend to involve relatively dense audio-visual activity throughout the clip, so the answer-bearing tokens are likely distributed across many regions and the diverse medoids recovered from the residual pool become more useful. In contrast, on the remaining benchmarks the discriminative content appears to be more localized within a smaller number of salient segments, so the top-scoring tokens already capture most of the relevant signal and clustering contributes minimally. Overall, these results suggest that the benefit of diversity-preserving clustering tends to grow with both audio-visual density and pruning aggressiveness.

\begin{figure}[t]
\centering
\includegraphics[width=\linewidth]{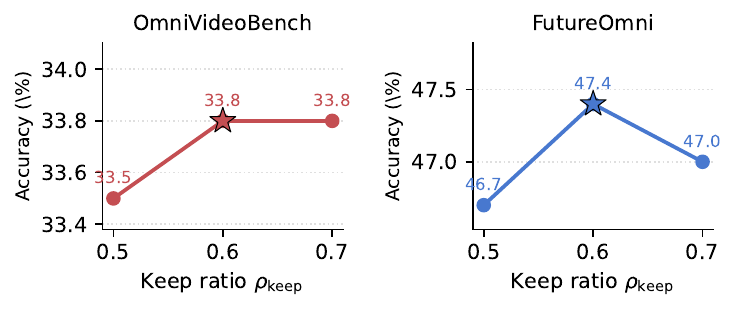}
\caption{\textbf{Ablation study of the per-window keep ratio $\rho_{\text{keep}}$.} We evaluate $\rho_{\text{keep}}$ on OmniVideoBench and FutureOmni to test its sensitivity, using Qwen2.5-Omni-7B at the $45\%$ retention budget.}
\label{fig:rho_keep}
\end{figure}

\paragraph{Ablation Study on $\rho_{\text{keep}}$.} Figure~\ref{fig:rho_keep} plots accuracy against the per-window keep ratio $\rho_{\text{keep}}$, which controls how much of the budget is committed by the score-based top-$k$ stage before cross-modal coupling. The curve is mildly concave and peaks at $\rho_{\text{keep}}{=}0.6$ on both benchmarks. Lowering it to $0.5$ starves the scoring stage and forces coupling to fill from less reliable candidates, while raising it to $0.7$ leaves less budget for pairing and drops pairs that coupling would have retained. This confirms that $\rho_{\text{keep}}{=}0.6$ best balances independent scoring and joint pairing. Moreover, accuracy stays within one point, indicating robustness to $\rho_{\text{keep}}$. More Experiments can be found in Appendix A.

\section{Conclusion}
\label{sec:conclusion}

This paper presents Omni-Prune, a novel plug-and-play framework that jointly prunes audio-video tokens for efficient OmniLLMs without requiring any training. Specifically, the framework partitions the interleaved token sequence at audio saliency peaks to form adaptive temporal windows, evaluates each token by combining encoder attention with text-query relevance, and binds matching audio-video tokens to preserve cross-modal consistency. As far as we are aware, this is the first pruning method designed for OmniLLMs that exploits all three modalities in the token selection process. Extensive experiments across diverse omnimodal understanding benchmarks with two model scales (3B, 7B) demonstrate that Omni-Prune consistently surpasses prior baselines. Notably, our method achieves significant memory reduction and prefill speedup while maintaining comparable accuracy.

\section*{Limitations}

\paragraph{Evaluation Scope.} Our study focuses on omnimodal video question answering across five widely used benchmarks spanning general-purpose, audio-centric, and long-form scenarios. While this protocol is sufficient to characterize the accuracy-efficiency profile of Omni-Prune, it does not exhaust the downstream tasks supported by omnimodal LLMs. Other tasks such as open-ended captioning and multi-turn dialogue may respond differently to token reduction, and their evaluation typically relies on human judgment rather than automatic metrics, which we leave to future work.

\paragraph{Hyperparameter Configuration.} Omni-Prune exposes a few control knobs, including the text-relevance floor $\tau$, the residual cluster ratio $\rho_c$, and the minimum window gap $f_{\min}$. To keep comparisons clean and demonstrate out-of-the-box usability, we adopt a single configuration across all benchmarks and both backbone scales. This setting is necessarily a compromise, since clip length and audio richness vary substantially across datasets, and lightweight per-dataset or adaptive tuning could yield further accuracy-efficiency gains.

\bibliography{references}

\appendix
\section{More Experimental Results}
\label{sec:appendix_more_results}

This section provides two additional ablations that complement the main paper: an extended sweep of the per-window keep ratio $\rho_{\text{keep}}$, and a transfer of the diversity-preserving clustering ablation to the smaller Qwen2.5-Omni-3B backbone.

\begin{table}[ht]
\centering
\small
\caption{\textbf{Extended ablation of $\rho_{\text{keep}}$ on Qwen2.5-Omni-7B at the $45\%$ budget.} We extend the sweep in Figure~\ref{fig:rho_keep} with two more extreme settings on OmniVideoBench and FutureOmni.}
\label{tab:rho_keep_ext}
\begin{tabular}{ccc}
\toprule
$\rho_{\text{keep}}$ & OmniVideoBench & FutureOmni \\
\midrule
0.4 & 33.4 & 46.1 \\
\rowcolor{oursrow}
0.6 (default) & 33.8 & 47.4 \\
0.8 & 33.9 & 47.1 \\
\bottomrule
\end{tabular}
\end{table}

\paragraph{Extended Ablation on $\rho_{\text{keep}}$.} Table~\ref{tab:rho_keep_ext} extends the main-paper sweep of $\rho_{\text{keep}}$ to the two more extreme settings $0.4$ and $0.8$. Pushing $\rho_{\text{keep}}$ down to $0.4$ leaves too little budget for the scoring stage and triggers visible drops on both benchmarks, whereas raising it to $0.8$ stays close to the peak at $0.6$ but slightly under-allocates to cross-modal pairing. The full $0.4$--$0.8$ range remains within roughly one accuracy point of the default, reinforcing the observation that Omni-Prune is robust to the exact choice of $\rho_{\text{keep}}$ as long as scoring and pairing both receive a non-trivial share of the budget.

\begin{table}[ht]
\centering
\small
\setlength{\tabcolsep}{3.5pt}
\caption{\textbf{Effect of diversity-preserving clustering on Qwen2.5-Omni-3B.} We compare Omni-Prune with and without the residual K-medoids step at the $45\%$ and $35\%$ budgets, and report per-benchmark scores.}
\label{tab:ablation_cluster_3b}
\resizebox{\linewidth}{!}{%
\begin{tabular}{lcccc}
\toprule
Variant & Daily-Omni & WorldSense & OmniV. & FutureO. \\
\midrule
\rowcolor{oursrow}
Full ($45\%$) & 58.8 & 45.9 & 32.6 & 47.0 \\
w/o Clustering & 58.7~\textcolor{red}{\scriptsize($-$0.1)} & 45.8~\textcolor{red}{\scriptsize($-$0.1)} & 32.5~\textcolor{red}{\scriptsize($-$0.1)} & 47.0~\textcolor{red}{\scriptsize($-$0.0)} \\
\midrule
\rowcolor{oursrow}
Full ($35\%$) & 58.5 & 45.4 & 31.6 & 46.9 \\
w/o Clustering & 57.5~\textcolor{red}{\scriptsize($-$1.0)} & 45.0~\textcolor{red}{\scriptsize($-$0.4)} & 31.3~\textcolor{red}{\scriptsize($-$0.3)} & 46.8~\textcolor{red}{\scriptsize($-$0.1)} \\
\bottomrule
\end{tabular}%
}
\end{table}

\paragraph{Diversity-Preserving Clustering on Qwen2.5-Omni-3B.} Table~\ref{tab:ablation_cluster_3b} mirrors the 7B clustering ablation in the main paper on the smaller 3B backbone. The trend observed at the 7B scale carries over: at the looser $45\%$ budget the drops from removing clustering are uniformly mild, whereas at the tighter $35\%$ budget the gap widens, most noticeably on Daily-Omni. This is consistent with the picture in the main paper.

\section{Related Work}
\label{sec:appendix_related}

This section expands Section~\ref{sec:related} with a treatment of audio-visual representation learning.

\paragraph{Audio-Visual Representation Learning.} Joint audio-visual learning predates OmniLLMs. AV-HuBERT~\cite{shi2022avhubert} pre-trains a unified audio-visual transformer via masked prediction; Video-LLaMA~\cite{zhang2023videollama} bridges audio and visual encoders to an LLM via Q-Formers; SALMONN~\cite{tang2023salmonn} and Video-SALMoNN~\cite{sun2024videosalmonn} extend LLMs with auditory perception including non-speech sounds; CLIP-style cross-modal alignment~\cite{radford2021clip} underlies many vision-language encoders used today.

\begin{figure*}[t]
\centering
\includegraphics[width=\linewidth]{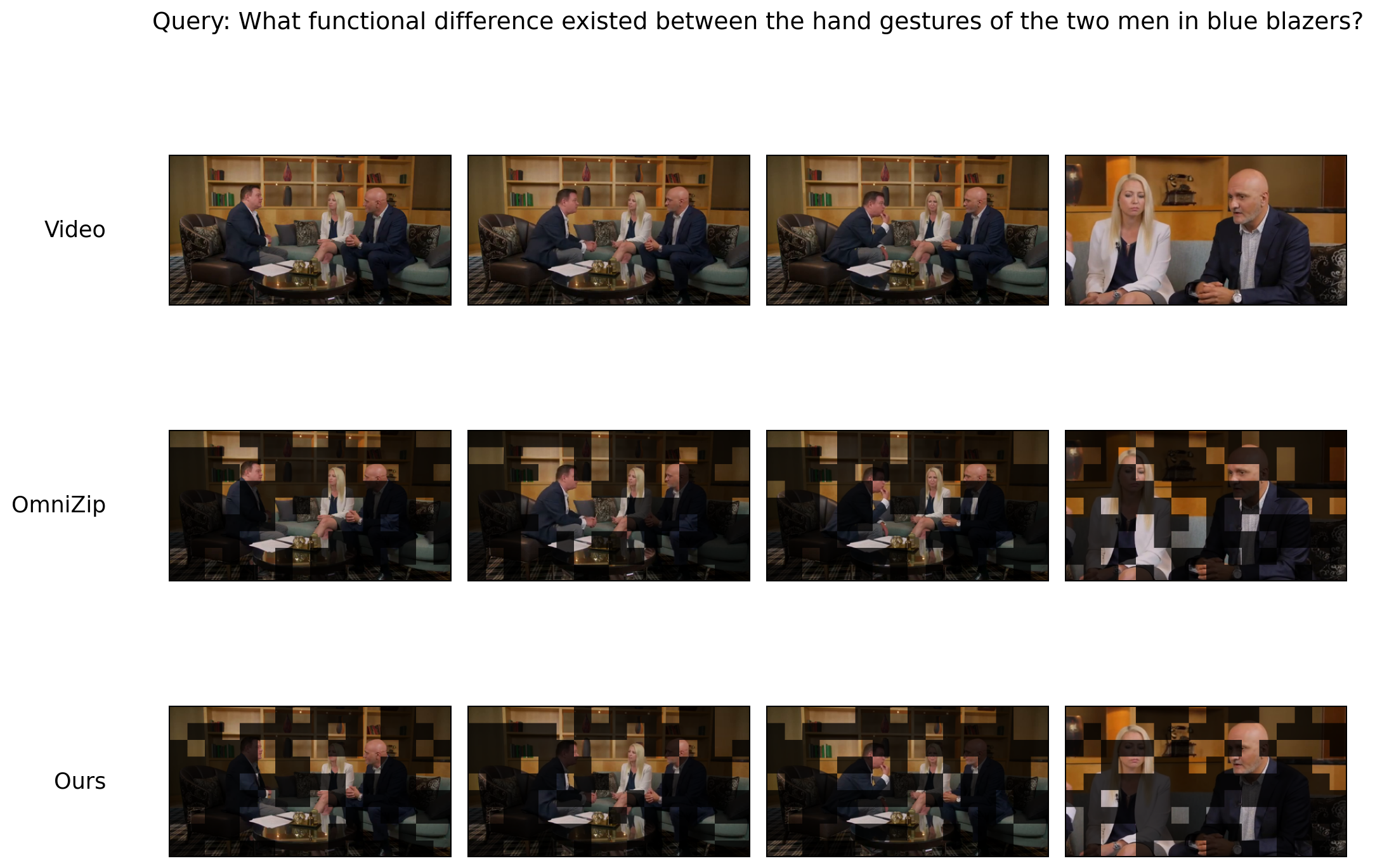}
\caption{Showcases of Omni-Prune \emph{vs.}~OmniZip with Qwen2.5-Omni 7B on a video input case.}
\label{fig:case_study}
\end{figure*}

\section{Notation and Implementation Details}
\label{sec:appendix_details}

Table~\ref{tab:notation} summarizes the notation used by Omni-Prune. The appendix follows the same formatting as Section~\ref{sec:method}: bold symbols denote vectors or matrices, calligraphic symbols denote token sets, and superscript $w$ denotes quantities restricted to a window.

\begin{table}[ht]
\centering
\small
\caption{Notation summary for Omni-Prune.}
\label{tab:notation}
\resizebox{\linewidth}{!}{%
\begin{tabular}{ll}
\toprule
\textbf{Symbol} & \textbf{Meaning} \\
\midrule
$\mathbf{X}$ & Interleaved audio-video-text token sequence \\
$\mathcal{A},\mathcal{V},\mathcal{T}$ & Audio, video, and text token index sets \\
$\mathbf{a}_i,\mathbf{v}_j,\mathbf{t}_j$ & Audio, video, and text token embeddings \\
$\mathbf{s}^a,\mathbf{s}^v$ & Encoder attention scores for audio/video tokens \\
$F,\,E_f,\,\tilde{E}_f$ & Frame count and raw/smoothed per-frame audio energy \\
$w,\mathcal{W}$ & Adaptive window and the full window set \\
$\mathcal{A}^w,\mathcal{V}^w$ & Audio/video tokens inside window $w$ \\
$v_i$ & Jointly normalized encoder saliency \\
$r_i^{\text{text}}$ & Text-query relevance score \\
$\phi_i$ & Composite retention score \\
$\mathbf{C}^w$ & Audio-video cosine relation map in window $w$ \\
$\mathcal{K}^w,\mathcal{P}^w$ & Per-window keep set and cross-modal paired set \\
$\mathbf{m}$ & Binary retention mask over all tokens \\
$\alpha,\rho_{\text{keep}},\rho_c$ & Global keep ratio, per-window keep cap, cluster ratio \\
$\tau,\gamma$ & Text-relevance floor and cross-modal pairing gate \\
$B,\,B_{\text{keep}}^w$ & Global and per-window keep budgets \\
\bottomrule
\end{tabular}%
}
\end{table}

\paragraph{Attention Score Extraction.} We compute $\mathbf{s}^a$ and $\mathbf{s}^v$ from the last encoder self-attention layer, following the definition in Section~\ref{sec:relation}. Scores are averaged across heads and accumulated over query positions, then assigned back to their corresponding audio or video tokens before the two modalities are jointly normalized inside each window.

\paragraph{Mask Construction.} The binary mask $\mathbf{m}$ spans the entire token sequence $\mathbf{X}$, but pruning is applied only to audio and video tokens: text tokens are always retained ($m_i=1$ for all $i\in\mathcal{T}$). After mask determination, the original token order in $\mathbf{X}$ is preserved when constructing the pruned sequence $\mathbf{X}'=\mathbf{X}[\mathbf{m}]$. This makes Omni-Prune plug-and-play for autoregressive OmniLLM decoders, as it does not alter token embeddings, positional indexing, or model parameters.

\paragraph{Budget Allocation.} The global audio-video budget is $B=\lfloor\alpha(|\mathcal{A}|+|\mathcal{V}|)\rceil$. Within each window, $B_{\text{keep}}^w$ caps the high-confidence keep set so that the subsequent cross-modal coupling stage still has capacity to retain paired evidence.

\section{Case Study}
\label{sec:appendix_casestudy}

Figure~\ref{fig:case_study} shows how query-aware scoring directs the budget toward the answer-bearing region. Guided by text relevance, Omni-Prune concentrates tokens on the interviewers' torsos and hands, and the cross-modal coupling stage pairs each gesture-bearing token with the synchronized speech. OmniZip relies primarily on audio attention, which leads it to retain more of the surrounding scenery.

\begin{figure*}[!htbp]
\centering
\begin{subfigure}{\linewidth}
  \centering
  \includegraphics[width=\linewidth]{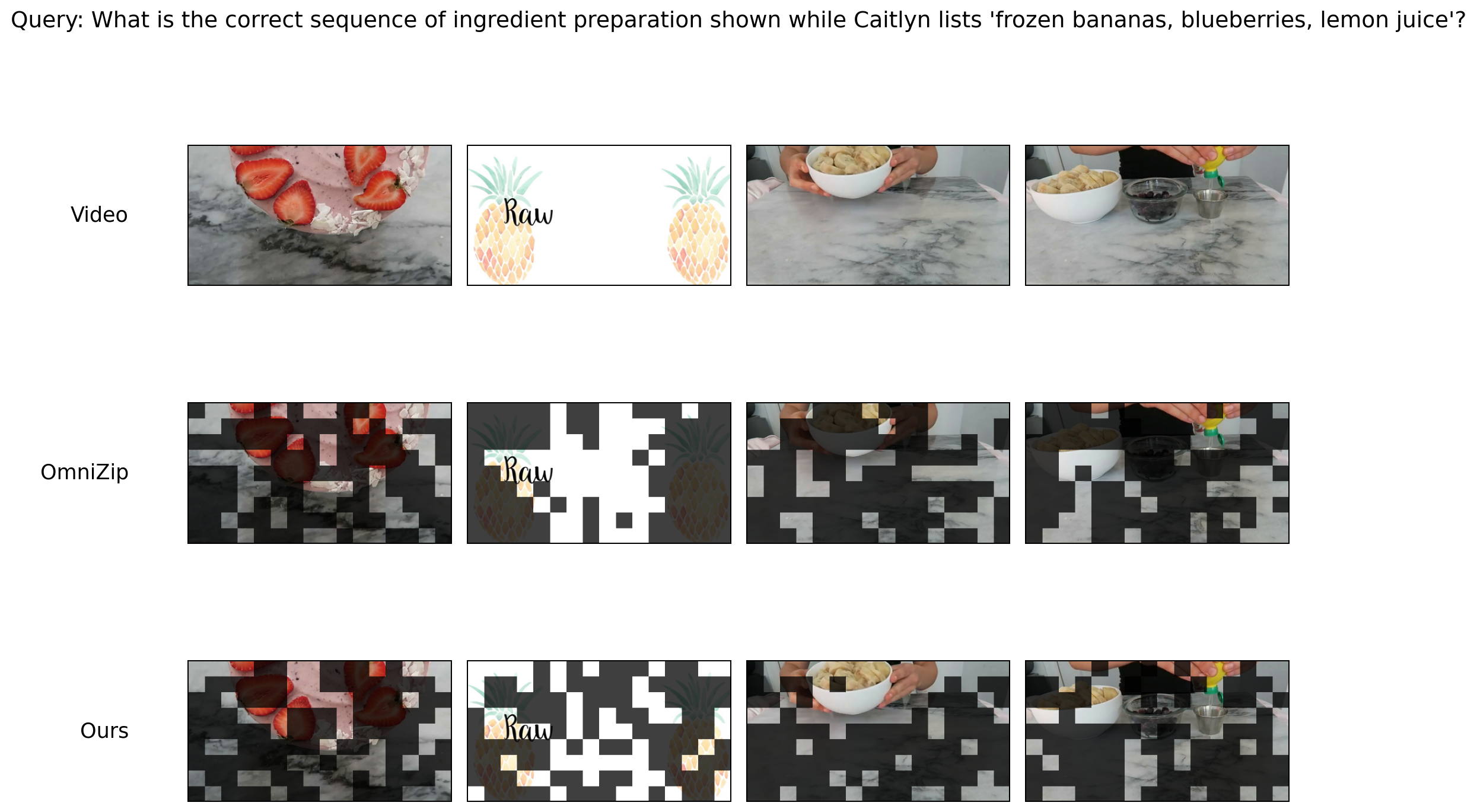}
  \phantomcaption
  \label{fig:case_ingredients}
\end{subfigure}\\[4pt]
\begin{subfigure}{\linewidth}
  \centering
  \includegraphics[width=\linewidth]{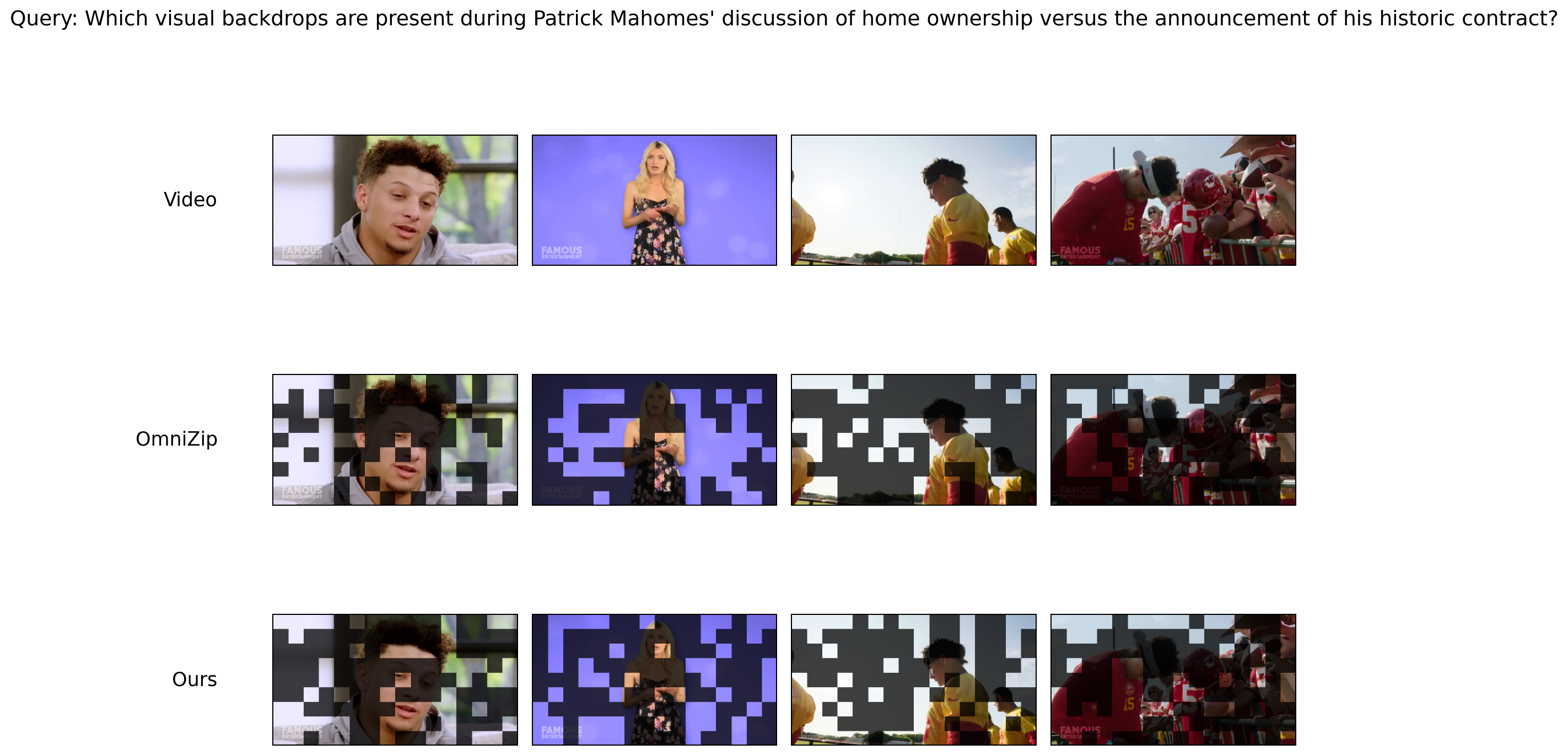}
  \phantomcaption
  \label{fig:case_mahomes}
\end{subfigure}
\caption{Showcases of Omni-Prune \emph{vs.}~OmniZip with Qwen2.5-Omni 7B on two additional video inputs.}
\label{fig:case_studies_extra}
\end{figure*}

Figure~\ref{fig:case_ingredients} illustrates how Omni-Prune tracks the entities named in the question. Tokens are reallocated toward the banana bowl, blueberry cup and lemon-squeezing hand, as text relevance aligns ingredient nouns with the corresponding patches and cross-modal pairing anchors them to the synchronized narration, so all three pieces of evidence survive the global selection step.

Figure~\ref{fig:case_mahomes} shows that, when the answer lies in the background, the same mechanism shifts the budget there. While maintaining comparable coverage of the foreground speakers, Omni-Prune additionally retains the window panes, purple studio gradient and stadium sky, because backdrop-related terms in the query are matched to the corresponding background patches through text relevance.

\section{Overall Algorithm}
\label{sec:appendix_algorithm}

The complete Omni-Prune pipeline consists of three stages.
We present the first two stages as separate algorithms below. Unless otherwise specified, all symbols follow Section~\ref{sec:method}: $w$ denotes an adaptive temporal window, $\mathcal{W}=\{w_1,\ldots,w_K\}$ denotes the set of windows, and $\mathcal{A}^w$/$\mathcal{V}^w$ denote the audio/video tokens inside window $w$.

\subsection{Stage 1: Audio-Anchor Adaptive Windowing}

Algorithm~\ref{alg:stage1} segments the token sequence into content-aware temporal windows using audio saliency peaks.

\begin{algorithm}[ht]
\caption{Audio-Anchor Adaptive Windowing}
\label{alg:stage1}
\small
\setlength{\algorithmicindent}{1.0em}
\begin{algorithmic}[1]
\REQUIRE Audio scores $\mathbf{s}^a$, frames $F$, gap $f_{\min}$
\ENSURE Windows $\mathcal{W}=\{w_1, \ldots, w_K\}$
\STATE $E_f \gets |T_f|^{-1}\sum_{i \in T_f} s^a_i$
\STATE $\tilde{\mathbf{E}} \gets \texttt{AvgPool1D}(\mathbf{E}, k{=}3)$
\STATE $\mathcal{P} \gets \emptyset$
\FOR{$f=2$ \TO $F-1$}
    \IF{$\tilde{E}_f \geq \tilde{E}_{f-1}$ \AND $\tilde{E}_f \geq \tilde{E}_{f+1}$}
        \STATE $\mathcal{P}.\texttt{append}(f)$
    \ENDIF
\ENDFOR
\STATE $\mathcal{B} \gets [0]$;\; $\text{last} \gets 0$
\FOR{$p \in \mathcal{P}$}
    \IF{$p - \text{last} \geq f_{\min}$}
        \STATE $\mathcal{B}.\texttt{append}(p)$;\; $\text{last} \gets p$
    \ENDIF
\ENDFOR
\IF{$\mathcal{B}[-1] \neq F$}
    \STATE $\mathcal{B}.\texttt{append}(F)$
\ENDIF
\STATE Derive $\mathcal{W}=\{w_1, \ldots, w_K\}$ from $\mathcal{B}$
\RETURN $\mathcal{W}$
\end{algorithmic}
\end{algorithm}

\subsection{Stage 2: Cross-Modal Relation Map and Budget Allocation}

Algorithm~\ref{alg:stage2} scores tokens within each window and allocates the global budget.

\begin{algorithm}[ht]
\caption{Cross-Modal Relation Map}
\label{alg:stage2}
\small
\setlength{\algorithmicindent}{1.0em}
\begin{algorithmic}[1]
\REQUIRE $\mathbf{X}$, $\mathcal{W}$, $\mathbf{s}^a$, $\mathbf{s}^v$, $\{\mathbf{t}_j\}$, $\alpha$, $\rho_{\text{keep}}$, $\tau$, $\gamma$
\ENSURE Binary mask $\mathbf{m}$
\STATE $B \gets \lfloor \alpha \cdot (|\mathcal{A}| + |\mathcal{V}|) \rceil$
\FOR{each window $w \in \mathcal{W}$}
    \STATE $\mathcal{M}^w \gets \mathcal{A}^w \cup \mathcal{V}^w$
    \STATE $\mathbf{s}^w \gets [\mathbf{s}^a_{\mathcal{A}^w};\; \mathbf{s}^v_{\mathcal{V}^w}]$
    \STATE $d^w \gets \max(\mathbf{s}^w) - \min(\mathbf{s}^w) + \epsilon$
    \STATE $v_i \gets (s^w_i - \min(\mathbf{s}^w))/d^w$
    \STATE $c_{ij} \gets \mathbf{x}_i^\top\mathbf{t}_j/(\|\mathbf{x}_i\|\,\|\mathbf{t}_j\|)$
    \STATE $r_i^{\text{text}} \gets \bigl(\max_j c_{ij} + 1\bigr)/2$
    \STATE $\phi_i \gets v_i + r_i^{\text{text}}$
    \STATE $\eta^w \gets \max(\bar{r}^{\text{text}}, \tau)$
    \STATE $\mathcal{Q}^w \gets \{i \in \mathcal{M}^w : v_i > \bar{v}^w\}$
    \STATE $\mathcal{Q}^w \gets \mathcal{Q}^w \cup \{i \in \mathcal{M}^w : r_i^{\text{text}} > \eta^w\}$
    \STATE $B_{\text{keep}}^w \gets \lfloor \alpha \rho_{\text{keep}} (|\mathcal{A}^w|+|\mathcal{V}^w|) \rceil$
    \STATE $\mathcal{K}^w \gets \texttt{topk}(\mathcal{Q}^w, B_{\text{keep}}^w, \text{key}{=}\phi_i)$
    \STATE $C^w_{ij} \gets \mathbf{a}_i^\top\mathbf{v}_j/(\|\mathbf{a}_i\|\,\|\mathbf{v}_j\|)$
    \STATE $\mathcal{P}^w \gets \texttt{bi\_greedy\_match}(\mathbf{C}^w, \gamma)$
\ENDFOR
\STATE $\text{kept} \gets 0$
\STATE $\mathcal{K} \gets \texttt{sorted}(\bigcup_{w\in\mathcal{W}} \mathcal{K}^w, \text{desc})$
\FOR{$(\phi_i, i) \in \mathcal{K}$}
    \IF{$\text{kept} < B$}
        \STATE $m_i \gets 1$;\; $\text{kept} \mathrel{+}{=} 1$
    \ENDIF
\ENDFOR
\STATE $\mathcal{P} \gets \texttt{sorted}(\bigcup_{w\in\mathcal{W}} \mathcal{P}^w, \text{desc})$
\FOR{$(s, i, j) \in \mathcal{P}$}
    \IF{$m_i=0$ \AND $m_j=0$ \AND $\text{kept} + 2 \leq B$}
        \STATE $m_i, m_j \gets 1$;\; $\text{kept} \mathrel{+}{=} 2$
    \ENDIF
\ENDFOR
\STATE Fill remaining $B - \text{kept}$ slots by $\phi_i$, minority modality first
\RETURN $\mathbf{m}$
\end{algorithmic}
\end{algorithm}

\section{Discussion}
\label{sec:appendix_discussion}

This section further discusses two key design choices in Omni-Prune: audio anchoring and the text-relevance floor.

\paragraph{Audio as the Temporal Anchor.} In OmniLLMs, the audio stream is typically much shorter than the video stream after tokenization yet carries strong temporal cues such as speech onsets, sound events, and silence boundaries. Using audio saliency to drive window partitioning therefore offers two practical advantages. First, audio peaks tend to align with semantic transitions in the video, so windows produced by audio-anchor segmentation rarely cut across coherent events. Second, because the audio sequence is short, peak detection and minimum-gap filtering only cost $O(F)$ and contribute negligible overhead to the overall prefill pipeline.

\paragraph{Role of the Text-Relevance Floor.} The composite score $\phi_i = v_i + r_i^{\text{text}}$ mixes encoder saliency and text-query relevance. Without a floor, the per-window average $\bar{r}^{\text{text}}$ would dominate when the question is generic (\eg ``what happens in this video?''), causing almost all tokens to look relevant and degenerating the selection into a pure saliency ranking. The floor $\tau{=}0.55$ ensures that text relevance only fires when a token is genuinely close to the query, which is particularly important for fine-grained question answering on benchmarks such as OmniVideoBench and FutureOmni.

\end{document}